\documentclass[conference]{IEEEtran}
\IEEEoverridecommandlockouts
\usepackage{cite}
\usepackage{amsmath,amssymb,amsfonts}
\usepackage{algorithmic}
\usepackage{graphicx}

\usepackage{textcomp}
\usepackage{xcolor}
\def\BibTeX{{\rm B\kern-.05em{\sc i\kern-.025em b}\kern-.08em
    T\kern-.1667em\lower.7ex\hbox{E}\kern-.125emX}}
\usepackage{tabularx}
\usepackage{booktabs}
\usepackage{float}
\DeclareUnicodeCharacter{2265}
{$\geq$}
\usepackage{array}
 
\usepackage{placeins}  % Garante o posicionamento de figuras e tabelas

\usepackage{booktabs}  % Para tabelas mais elegantes
\usepackage{multicol}  % Para colunas adicionais em tabelas
\usepackage{multirow}  % Para células mescladas em tabelas
\usepackage{placeins}  % Forçar a ordem de floats (ex.: \FloatBarrier)
\usepackage[font=small,skip=0pt]{caption}

\usepackage{array}  % Adicionar no preâmbulo
\usepackage{adjustbox} % Ajustar a tabela dentro do espaço

\author{
{\rm Giuliano Lorenzoni}\\
glorenzo@uwaterloo.ca\\
University of Waterloo\\
Waterloo, Ontario, Canada
\and
{\rm Pedro Elkind Velmovitsky}\\
pedro.velmovitsky@uhn.ca\\
Centre for Digital Therapeutics\\
Toronto, Ontario, Canada

\and
{\rm Paulo Alencar}\\
palencar@uwaterloo.ca\\
University of Waterloo \\
Waterloo, Ontario, Canada

\and
{\rm Donald Cowan}\\
dcowan@uwaterloo.ca\\
University of Waterloo \\
Waterloo, Ontario, Canada
}

\begin{document}
\title{GPT-4 on Clinic Depression Assessment: An LLM-Based Pilot Study}

\maketitle

\begin{abstract}
depression has impacted millions of people worldwide and has become one of the most prevalent mental disorders. Early mental disorder detection can lead to cost savings for public health agencies and avoid the onset of other major comorbidities. Additionally, the shortage of specialized personnel is a critical issue because clinical depression diagnosis is highly dependent on expert professionals and is time-consuming. 

In this study, we explore the use of GPT-4 for clinical depression assessment based on transcript analysis. We examine the model's ability to classify patient interviews into binary categories: depressed and not depressed. A comparative analysis is conducted considering prompt complexity (e.g., using both simple and complex prompts), as well as varied temperature settings, to assess the impact of prompt complexity and randomness on the model's performance. Results indicate that GPT-4 exhibits considerable variability in accuracy and F1-Score across configurations, with optimal performance observed at lower temperature values (0.0–0.2) for complex prompts. However, beyond a certain threshold (temperature ≥ 0.3), the relationship between randomness and performance becomes unpredictable, diminishing the gains from prompt complexity. These findings suggest that, while GPT-4 shows promise for clinical assessment, the configuration of the prompts and model parameters requires careful calibration to ensure consistent results. This preliminary study contributes to understanding the dynamics between prompt engineering and large language models, offering insights for future development of AI-powered tools in clinical settings.

\end{abstract}

\begin{IEEEkeywords}
 LLM, GPT, Clinical Depression,classification models, Variability
\end{IEEEkeywords}

\section{Introduction}

Depression affects millions of people worldwide and is recognized as one of the most prevalent mental disorders. Early detection of mental health disorders can significantly reduce healthcare costs and prevent the development of severe comorbidities. However, the shortage of specialized professionals required for precise diagnoses is an increasing concern, as the process heavily relies on experts and demands considerable time.

In this study, we propose an innovative approach to clinical depression assessment by leveraging GPT-4 as a diagnostic support tool. Through the analysis of interview transcripts, we aim to explore how artificial intelligence can replicate the clinical sensitivity and diagnostic abilities of human professionals. Beyond simple and direct prompts, we investigate the impact of more sophisticated prompt designs, enriched with detailed examples and fine-tuned temperature settings, to optimize the model's accuracy and consistency. This study not only evaluates GPT-4's capacity to identify depression cases but also explores how prompt structure and temperature calibration influence the stability and predictability of responses, setting new standards for the use of AI in mental health care.

To guide our investigation and deepen the analysis, we formulated the following research questions:

\begin{itemize}
    \item \textbf{RQ1}: Can Foundational LLMs accurately diagnose clinical depression using simple binary classification approaches?
    \item \textbf{RQ2}: Does few-shot prompting with examples of therapy sessions help foundational LLMs' accurately diagnose clinical depression?
    \item \textbf{RQ3}: Does more complex prompt engineering improve LLMs' ability to diagnose clinical depression?
    \item \textbf{RQ4}: What is the effect of temperature on LLMs' ability to diagnose depression?
    \item \textbf{RQ5:} How does output variability influence the reliability of GPT-4 in clinical depression diagnosis?
\end{itemize}

This paper is structured into six main sections. Section 2 presents a review of the relevant literature, discussing the state-of-the-art language models applied to mental health and clinical diagnostics. Section 3 outlines the experimental design, covering data usage and the prompt engineering strategies applied to GPT-4. Section 4 presents the results obtained under different prompt configurations and temperature settings, analyzing the model's performance for each approach. Section 5 discusses the findings, highlighting the role of randomness, dataset imbalance, and the sensitivity of evaluation metrics under varying temperature conditions. Finally, Section 6 provides the conclusions and suggests avenues for future research, including advancements with more sophisticated techniques such as RAG and fine-tuning, to enhance diagnostic accuracy and promote the adoption of AI in mental healthcare.

\section{Related Work} \label{sec:related}
% prompt

% model parameters/temperature 

% Novelty: In contrast, the proposed experimental study considers variabilities related to prompt complexity, level of randomness as measured by temperature, and performance metrics.

% We identified one closely related study focused on healthcare, which aligns directly with our research objectives. Additionally, we reviewed other relevant works from various domains that employ techniques and methodologies similar to those used in our study. 

Several studies have employed the DAIC-WOZ database for depression diagnosis, focusing on Machine Learning (ML) techniques rather than Large Language Models (LLMs). Some authors \cite{39yang2019depression} applied Support Vector Machines (SVMs) and TextCNN to text-based emotional analysis, exploring the potential of these models to detect linguistic patterns associated with depression. Other authors \cite{bhavya2022speech} investigated emotion recognition using speech patterns within the DAIC-WOZ dataset, leveraging acoustic features for depression detection. Similarly, one study \cite{cummins2018speech} used DAIC-WOZ, emphasizing the importance of acoustic features in building robust models for mental health assessments. While these studies utilize the DAIC-WOZ dataset, their focus on ML techniques distinguishes them from our work, which explores LLM-based classification for clinical purposes.

In addition to research using DAIC-WOZ, several other studies have explored ML techniques for depression diagnosis and sentiment analysis, leveraging different datasets and methodologies. These studies address challenges such as emotion analysis on social media, multimodal data integration, and enhanced prediction through feature selection and parameter tuning. Examples include the works described in \cite{liu2022detecting}
,\cite{squires2023deep}, \cite{krishna2020different}, \cite{nguyen2023multimodal}, \cite{mao2023systematic}, \cite{aleem2022machine}, \cite{ndaba2023review}, \cite{nouman2021recent}, \cite{da2022review}, and \cite{yan2022challenges}. 
Although these studies do not use the same dataset or employ LLMs, they offer complementary perspectives, showcasing the role of ML in mental health diagnostics and sentiment analysis.

Specifically, regarding LLM-based approaches, a few studies provide insights into how LLMs are being applied across fields, offering complementary perspectives for the improvement of our approach. % The study 'Exploring the Capabilities of a Language Model Only Approach for Depression Detection in Text Data'
 Some authors
\cite{sadeghi2023exploring} investigate the use of LLMs for depression detection through text analysis, using the E-DAIC dataset, an extension of the DAIC-WOZ dataset. The primary goal of this research is to predict PHQ-8 scores through automated techniques, replacing manual feature extraction. This study achieved a Mean Absolute Error (MAE) of 3.65, demonstrating the model's capability to evaluate depression severity effectively. Our work aligns with this study as we also employ the DAIC-WOZ dataset; however, our approach diverges by focusing on binary classification (depressed or not depressed). Additionally, we explore the impact of various prompt strategies and temperature calibration to optimize the stability and accuracy of the classification task. While this article is the most comparable to our research, we emphasize stability and performance optimization in clinical tasks, extending the potential of LLMs in healthcare.

In addition to healthcare, several studies from other fields employ methodologies and techniques relevant to our work.

% "Large Language Models: The Next Frontier for Variable Discovery within Metamorphic Testing?"

The study described in \cite{tsigkanos2023large} explores the use of LLMs to automate variable extraction from scientific software manuals, removing the need for manual intervention. Similar to our study, it emphasizes the importance of prompt design and parameter calibration using few-shot examples and temperature adjustments to enhance performance. Although our study also focuses on variable extraction for scientific testing, it applies LLMs in clinical classification, introducing unique challenges such as minimizing false negatives to avoid clinical risks. Both studies share the need to ensure accuracy and stability in the output.

% "Hybrid Prompt-Driven Large Language Model
% for Robust State-of-Charge Estimation
% of Multi-Type Li-ion Batteries" 
Othe authors 
\cite{bian2024hybrid} explore using LLMs to estimate the state of charge (SOC) of lithium-ion batteries. While the context differs, there are meaningful parallels between the two studies, as both utilize prompt-based strategies. Our focus lies on binary classification for depression diagnosis, whereas this article applies LLMs for numerical estimation in an industrial context. Moreover, the referenced study employs a hybrid approach with task-specific knowledge through soft prompts, offering potential inspiration for future healthcare applications by enabling the seamless integration of clinical data into prompts.

% The article "Evaluating Diverse Large Language Models for Automatic and General Bug Reproduction" 
Another study
\cite{kang2024evaluating} investigates the use of LLMs for automating test generation from bug reports through the LIBRO pipeline. This study compares 15 LLMs, assessing their effectiveness in bug reproduction through different parameters, including model size and temperature. Although in a distinct context, both studies rely on prompt strategies and temperature calibration to optimize performance. The difference lies in the focus: our study addresses clinical classification, while theirs explores software bug reproduction. However, both emphasize the importance of consistent predictions and temperature tuning, showing that insights from one domain can be valuable in another.

Finally, another relevant study explores the use of GPT-3.5 and GPT-4 to generate automated annotations for emotion, sentiment, and cognitive dissonance in financial conference calls, comparing them to human-annotated data. 
% The study "Humans vs. ChatGPT: Evaluating Annotation
% Methods for Financial Corpora" 
This study
\cite{kaikaus2023humans} demonstrates that LLMs not only outperform human annotators in terms of consistency and reliability but also offer significant cost and time efficiencies. Additionally, it examines the impact of different prompt strategies on annotation quality, particularly the role of contextual information. While the referenced article focuses on emotions and sentiment analysis within the financial sector, our work applies binary classification to mental health, highlighting the adaptability of LLMs across diverse fields and the importance of tailoring prompt strategies to specific tasks.

In summary, while our study shares similarities with works from various domains, it uniquely contributes to the growing body of research on LLMs in mental health by focusing on binary classification for depression detection. Through prompt design and temperature calibration, we aim to optimize performance and stability in a clinical context. A key novelty of our experimental approach lies in its consideration of variabilities related to prompt complexity, the level of randomness as measured by temperature, and performance metrics. The insights from these related studies, although applied in different fields, provide valuable perspectives that can inspire future improvements in healthcare-focused LLM applications.

\section{Experiment Design}

This study leverages collected data exclusively for testing and evaluation, without any training phase, ensuring the validity of results. The approach relies on carefully designed prompts and their interaction with GPT-4's classification capabilities, utilizing responses generated via the OpenAI API. Key examples from the dataset, including two positive and two negative cases, were selected to enhance the prompts. The prompt engineering process involved four stages: (1) A simple binary classification using GPT-4 with a basic prompt (“depressed” or “not depressed”), (2) Improving precision by integrating four representative examples, (3) Refining the approach with an elaborate prompt that adds clinical context alongside the examples, and (4) Exploring the impact of temperature calibration (0.0, 0.1, 0.2, etc.) to optimize accuracy and F1-Score. Computationally, the study utilized the OpenAI API to access GPT-4, with Pandas and NumPy for data management, scikit-learn for performance metrics calculation, and Matplotlib/Seaborn for visualizing the outcomes.

% The objective of this case study is to build a diagnostic model for depression disorders based on different supervised ML models and NLP techniques. We explored multiple model tuning configurations, feature sets, and data preprocessing methodologies across all the models incorporated in our analysis. 

% As a result, the design of this case study was conceived to identify the optimal model, the most suitable parameters, and any combination of factors that could provide the best results in terms of accuracy. We also established a baseline approach to check the efficiency of the models and examine the insights of our case study based on our results.

\subsection{Data Collection: Dataset}\label{AA}
% Explain the process for finding a database to support our research
We based our research on the Distress Analysis Interview Corpus - Wizard-of-Oz (DAIC-WOZ), a dataset designed to support the diagnosis of mental disorders such as depression, anxiety, and post-traumatic stress disorder. DAIC is a database that is part of a larger corpus \cite{10gratch2014distress} %[10]  
available to the research community by request. We have submitted a request and the data released refers only to the depressed patients' database.

This database contains clinical interviews conducted by humans, human-controlled agents, and autonomous agents. The computer agent is an animated virtual interviewer robot called Ellie that identifies mental illness indicators. Data includes 189 sessions of interviews which correspond to questionnaire responses and audio and video recordings. Each interview is identified by a session number and has a correspondent folder of files which includes files related to video features, audio, transcript, and audio features. This dataset contains training, development, and test subsets. Interviewees are both distressed and non-distressed individuals. 

We used the text files related to the transcript of the interviews. The Patient Health Questionnaire depression scale (PHQ-8) defines the depression diagnostic and severity measure.  and the PHQ-8 file which described the score of each patient according to the depression scale described in  \cite{11kroenke2009phq}.%[11]% \cite{kroenke2009phq}. 

\subsection {Prompt Engineering Process}

This section outlines the different strategies employed to develop and refine prompts for evaluating the GPT-4 model’s ability to classify between "depressed" and "not depressed." Progressively more sophisticated approaches were explored, culminating in temperature calibration to optimize accuracy and response stability.

\subsubsection{RQ1 Method - Simple Prompt}
In this initial approach, a basic prompt was used, asking GPT-4 to provide a binary classification between "depressed" and "not depressed" for each transcript. The goal of this stage was to assess the model’s performance with a straightforward instruction, without any additional context. This method serves as a baseline, offering a reference point for comparison with more advanced approaches.

\subsubsection{RQ2 Method - Prompt with Examples}
To improve the model's accuracy, four examples from the dataset were introduced: two positive examples (classified as "depressed") and two negative examples (classified as "not depressed"). 

The four examples were selected to provide a baseline of interview cases for each category without introducing additional criteria. The primary goal was to ensure the examples resembled other interviews of the same classification to avoid overfitting or introducing artifacts unrelated to the classification task. The choice to use four examples aimed at balancing the need for context enrichment with minimizing the reduction in available observations for analysis. While the distribution of examples was balanced at 50\% per class, this slightly deviates from the original dataset distribution of approximately 30\% "depressed" cases. However, care was taken to ensure the overall proportion of test data remained representative of the dataset.

These examples were incorporated into the prompt to provide GPT-4 with additional context regarding the expected classification. The inclusion of these examples had a significant impact, enhancing the model's ability to generalize and refine its predictions.

\subsubsection{RQ3 Method - Detailed Prompt with Examples}
In this phase, the prompt was further refined to include more detailed instructions and a richer clinical context. In addition to the four examples used previously, the prompt was expanded to simulate the perspective of an expert in psychopathology, guiding the model to deliver a more robust and accurate analysis. This approach aimed to evaluate how the complexity of the prompt influences performance, especially for tasks requiring contextualized analysis.

\subsubsection{RQ4 Method - Effect of Temperature on LLM's Inferences}
To investigate the impact of temperature on the stability and performance of classifications, different temperature values (0.0, 0.1, 0.2, 0.3, and 0.5) were tested. 

\subsubsection{RQ5 Method - Temperature Calibration and Stability Analysis}
Temperature controls the degree of randomness in the model’s responses: lower values tend to produce more consistent responses, while higher values increase response diversity. The goal of this analysis was to explore the balance between stability and diversity, identifying the optimal setting to maximize the model's performance in complex classification tasks.

% \begin{itemize}
    
% \item Random Forest Classifier (RF): an algorithm that builds multiple decision trees and makes a final prediction based on the majority outcome of each decision tree.
% \item XGBoost (XGB): an algorithm that sequentially builds decision trees that recurrently diminish the error of the previous prediction by applying gradient descent on a loss function when adding new trees.
% \item Support Vector Machine (SVM): an algorithm that maps the data points to another space (using kernel algorithms) so that it is possible to separate (and therefore classify) the data points into different categories. 
% \end{itemize}

\subsection{Computational Packages}

The computational framework for this study was built using various Python libraries and tools to ensure an efficient and reproducible experimental pipeline. The following packages were utilized:

\begin{itemize}
    \item \textbf{OpenAI API}: Provides access to GPT-4, which was used for binary classification tasks in the context of clinical depression assessment. The API allowed dynamic prompt engineering, integration of contextual examples, and parameter tuning, such as temperature adjustments, to explore model performance.
    
    \item \textbf{Pandas and NumPy}: These libraries were essential for data manipulation and analysis. Pandas facilitated data cleaning, handling data structures such as DataFrames, and exporting the results to Excel files. NumPy was used to support mathematical operations throughout the experiment.
    
    \item \textbf{scikit-learn}: This library was employed to calculate key performance metrics such as Accuracy, Precision, Recall, and F1-Score. Additionally, it provided tools for generating confusion matrices, which were crucial for evaluating the model’s classification outcomes.
    
    \item \textbf{Matplotlib/Seaborn}: These libraries were used to create visualizations of the experimental results. Matplotlib allowed the plotting of line graphs to analyze the relationship between temperature settings and performance metrics. Seaborn complemented the analysis by generating heatmaps of confusion matrices, helping to identify patterns in the model’s classifications.
\end{itemize}

This combination of tools enabled seamless data processing and model evaluation, ensuring that the insights generated were both comprehensive and transparent. The modularity of the Python code allowed easy experimentation with different prompts and temperature settings, which was critical for refining the results.

\section{Results}
\subsection{RQ1: Baseline Classification with Simple Prompt}
This subsection presents the results of using a simple prompt to classify texts as "depressed" or "not depressed". 
The model achieved an accuracy of 70.74\%, but the recall was only 10.71\%, indicating that while the model successfully identified most negative cases, it struggled to correctly detect positive cases (depressed). 

\begin{table}[H]
\centering
\caption{Performance Metrics for Simple Prompt}
\begin{tabular}{lr}
\toprule
\textbf{Metric} & \textbf{Value} \\
\midrule
Accuracy  & 70.74\% \\
Precision & 54.55\% \\
Recall    & 10.71\% \\
F1-Score  & 17.91\% \\
\bottomrule
\end{tabular}
\end{table}

\begin{table}[H]
\centering
\caption{Confusion Matrix for Simple Prompt}
\begin{tabular}{cc}
\toprule
127 & 5 \\
50  & 6 \\
\bottomrule
\end{tabular}
\end{table}

In Table II, 188 out of 189 available observations were used, with one observation presenting structural issues, making its processing impossible. This result reflects the initial robustness of the pipeline for this specific configuration but also highlights limitations in handling problematic data.

The use of a simple prompt for classification presented significant limitations, as evidenced by the low F1-Score of 17.91\%. This is due to the model's inability to balance precision and recall. Although the accuracy is relatively high (70.74\%), the recall of only 10.71\% indicates that the model failed to correctly identify most positive depression cases, resulting in a high number of false negatives (50). This shortcoming is critical in clinical diagnostic applications, where sensitivity is essential to ensure that patients in need of intervention are not overlooked. The simplicity of the prompt may have led the model to miss important nuances present in the data, thus reducing its effectiveness for the intended task.

\subsection{RQ2: Performance Enhancement with Simple Prompt and Example-based Classification}
This approach introduces four examples in the prompt (two for each class), resulting in a significant improvement in recall to 77.78\% and an F1-Score of 60.87\%. However, the accuracy dropped slightly to 70.49\%, reflecting increased sensitivity to positive cases but at the cost of more false positives.

\begin{table}[H]
\centering
\caption{Performance Metrics for Example-based Prompt}
\begin{tabular}{lr}
\toprule
\textbf{Metric} & \textbf{Value} \\
\midrule
Accuracy  & 70.49\% \\
Precision & 50.00\% \\
Recall    & 77.78\% \\
F1-Score  & 60.87\% \\
\bottomrule
\end{tabular}
\end{table}

\begin{table}[H]
\centering
\caption{Confusion Matrix for Example-based Prompt}
\begin{tabular}{cc}
\toprule
87 & 42 \\
12 & 42 \\
\bottomrule
\end{tabular}
\end{table}

In Table IV, four examples were directly used in the prompt, reducing the number of available observations for analysis to 184. Of these, 183 were successfully processed, while one observation encountered issues during the pipeline. This suggests that, despite the improvements introduced by the examples, the model's sensitivity to certain data characteristics can still cause difficulties in specific cases.

Using the simple prompt with accompanying classification examples demonstrated significant improvements, as reflected in the substantial increase in recall to 77.78\% and the F1-Score. The inclusion of two examples for each class (depressed and not depressed) provided the model with clearer contextual references, enabling it to better capture the nuances required for accurate classification. This strategy reduced the number of false negatives, significantly improving the model’s sensitivity—a critical aspect for clinical applications. While the accuracy decreased slightly to 70.49\%, this trade-off is justified, as the enhanced recall ensures that a larger proportion of individuals with depression are correctly identified. The improved balance between precision and recall highlights the effectiveness of example-based prompts in guiding the model toward more reliable predictions.

\subsection{RQ3: Complex Prompt Design with Example-based Classification}
This configuration uses a more detailed prompt, including clinical context and the same classification examples. However, despite the added sophistication, the results showed that accuracy dropped to 69.23\% and the F1-Score was 51.72\%.

\begin{table}[H]
\centering
\caption{Performance Metrics for Complex Prompt with Examples}
\begin{tabular}{lr}
\toprule
\textbf{Metric} & \textbf{Value} \\
\midrule
Accuracy  & 69.23\% \\
Precision & 48.39\% \\
Recall    & 55.56\% \\
F1-Score  & 51.72\% \\
\bottomrule
\end{tabular}
\end{table}

\begin{table}[H]
\centering
\caption{Confusion Matrix for Complex Prompt with Examples}
\begin{tabular}{cc}
\toprule
96 & 32 \\
24 & 30 \\
\bottomrule
\end{tabular}
\end{table}

In Table VI, four examples were again incorporated, resulting in 184 available observations. In this case, two observations were not processed due to failures similar to those identified in the previous experiment. These inconsistencies can be attributed to the increased complexity of the prompts, which may have intensified the model's sensitivity to data variations.

The use of a complex prompt with classification examples, despite its added sophistication, yielded inferior results compared to the simpler prompt with examples. One plausible explanation for this degradation lies in the default temperature setting, which introduces randomness into the model’s output. Complex prompts may amplify this randomness, as the model processes additional contextual details that increase the likelihood of divergent interpretations. This effect is particularly noticeable in scenarios requiring precise classifications, where simpler prompts may reduce ambiguity and maintain focus on the core task. Exploring temperature calibration in RQ4 provides further insights into managing this variability.

These challenges are reflected in the observed metrics: the accuracy dropped to 69.23\%, and the F1-Score decreased to 51.72\%. This suggests that increasing the complexity of the prompt introduced unintended variability in the model's responses, possibly making the task more ambiguous for the GPT-4 model. While the complex prompt provided more detailed clinical context, this added information might have led the model to overfit specific details or deviate from the intended binary classification task. The lower recall of 55.56\% further indicates that the model struggled to identify a sufficient number of positive cases, diminishing its practical utility for clinical applications where high sensitivity is essential. Thus, the results highlight that more elaborate prompts may not always enhance performance and can even hinder the model’s consistency, particularly when clarity and precision are paramount.

\subsection{RQ4: Temperature Calibration and Stability with Complex Prompt and Examples}

\begin{table}[h!]
\centering
\caption{Performance Metrics Across Different Temperatures}
\label{tab:performance_metrics}
\resizebox{\columnwidth}{!}{%
\begin{tabular}{lcccc}
\toprule
\textbf{Temperature} & \textbf{Accuracy (\%)} & \textbf{Precision (\%)} & \textbf{Recall (\%)} & \textbf{F1-Score (\%)} \\ 
\midrule
0.0 & 72.28 & 51.95 & 74.07 & 61.07 \\ 
0.1 & 73.37 & 53.09 & 79.63 & 63.70 \\ 
0.2 & 71.74 & 51.16 & 81.48 & 62.86 \\ 
0.3 & 67.93 & 46.67 & 64.81 & 54.26 \\ 
0.5 & 68.48 & 47.56 & 72.22 & 57.35 \\ 
\bottomrule
\end{tabular}%
}
\end{table}

\begin{table}[ht]
\centering
\caption{Confusion Matrices for Different Temperature Settings}
\renewcommand{\arraystretch}{4.5}  % Ajusta a altura das linhas
\setlength{\tabcolsep}{12pt}  % Ajusta a largura entre colunas
\resizebox{\columnwidth}{!}{%
\begin{tabular}{|>{\Large}c|>{\Huge}c|>{\Huge}c|>{\Huge}c|>{\Huge}c|}
\hline
\textbf{\Large Temperature} & \textbf{\Large Actual ND | Predicted ND} & \textbf{\Large Actual D | Predicted ND} & \textbf{\Large Actual ND | Predicted D} & \textbf{\Large Actual D | Predicted D} \\ \hline
\textbf{\Large 0.0} & 93 & 14 & 37 & 40 \\ \hline
\textbf{\Large 0.1} & 92 & 11 & 38 & 43 \\ \hline
\textbf{\Large 0.2} & 88 & 10 & 42 & 44 \\ \hline
\textbf{\Large 0.3} & 90 & 19 & 40 & 35 \\ \hline
\textbf{\Large 0.5} & 87 & 15 & 43 & 39 \\ \hline
\end{tabular}%
}
\end{table}

In the temperature calibration experiments, the pipeline was enhanced to address the previously identified issues. These enhancements, combined with the robustness of the adjusted pipeline, allowed all 184 observations to be processed successfully, regardless of the temperature settings explored.

\begin{itemize}
    \item \textbf{Temperature 0.0:}  
    The model achieved 72.28\% accuracy and an F1-Score of 61.07\%, indicating increased predictability and consistency.

    \item \textbf{Temperature 0.1:}  
    Lowering the temperature to 0.1 further improved accuracy to 73.37\% and the F1-Score to 63.70\%.

    \item \textbf{Temperature 0.2:}  
    At 0.2, recall increased to 81.48\%, but accuracy dropped to 71.74\%.

    \item \textbf{Temperature 0.3:}  
    At 0.3, performance dropped significantly, with an F1-Score of 54.26\% and accuracy of 67.93\%.

    \item \textbf{Temperature 0.5:}  
    Finally, at 0.5, the model achieved 68.48\% accuracy and an F1-Score of 57.35\%.
\end{itemize}

The adjustments in temperature settings played a crucial role in optimizing the model's performance. Lower temperatures (0.0 to 0.2) provided greater predictability and stability, resulting in more consistent classifications, as evidenced by the gradual improvement in accuracy and F1-Score at temperatures 0.0 and 0.1.

\subsection{RQ5: Temperature Calibration and Stability Analysis}

The analysis of temperature calibration revealed critical insights into GPT-4's performance in sensitive clinical classification tasks. The use of structured prompts with illustrative examples proved effective, providing a solid foundation for achieving consistent metrics, particularly in terms of Accuracy and F1-Score. Stability in the results was primarily observed within the temperature range between 0.0 and 0.2, where the model's responses were predictable, minimizing unwanted variations and promoting reliable behavior.

Adjustments within this temperature range optimize the trade-off between predictability and performance, ensuring that the model’s classifications remain consistent, especially for high-stakes clinical contexts such as depression detection. As the temperature increased beyond 0.3, we observed a marked rise in output randomness, compromising the model’s precision and leading to more inconsistent classifications.

This behavior suggests the existence of an optimal temperature range for complex prompts with clinical examples. Lower temperatures (between 0.0 and 0.2) offer the best balance between stability and sensitivity, whereas higher temperatures introduce unpredictability, making the model less suitable for clinical use. Proper calibration of this parameter is therefore essential to balance the model’s inherent randomness with the need for stability in clinical settings. This balance is crucial to ensure that the model can provide reliable responses aligned with healthcare professionals’ expectations, minimizing risks associated with false diagnoses or inconsistent classifications.

\section{Discussion}

\subsection{Effect of Default Randomness on Results}
In our experiments, we observed that even with the same parameter configuration, results varied due to the presence of random components in the model. This variation can influence both accuracy and F1-Score, particularly across different runs with the same temperature and prompt. A potential solution to mitigate this effect would be to calculate the average of results over several runs using the same configuration. This approach would reduce the impact of randomness, providing a more robust and consistent evaluation of the model’s performance.

\subsection{Unbalanced Dataset and Impact on Accuracy}
The dataset used in this study is significantly unbalanced, with 56 out of 188 interviews showing symptoms of depression, while the rest represent subjects without depression. This uneven distribution suggests that the F1-Score is a particularly relevant metric, as it balances precision and recall, better reflecting the model’s ability to identify both classes. It is worth noting that the dataset exhibits a disproportionately high ratio of depressed cases (56 out of 188, approximately 30\%), which is significantly higher than the estimated global prevalence of depression in the general population. This imbalance would create a bias toward the classification of subjects as "depressed," potentially inflating recall metrics at the cost of generalizability to real-world scenarios.
 However, since the study does not involve a training process but instead performs direct classification on the dataset, accuracy remains an important metric. 

% Another important consideration involves the use of examples to enhance the prompt. The inclusion of two examples for each class (50\% "depressed" and 50\% "not depressed") introduces a slight imbalance compared to the real distribution in the dataset. Although this strategy improved recall, uncertainties remain about how the artificially balanced examples affect overall performance and the model’s generalization capabilities. \textcolor{red}{Additionally, we don't know if the artificially increased proportion of depressed cases within the dataset amplifies the potential bias introduced by the balanced examples used in the prompts. While this approach improved recall, it may limit the model’s ability to generalize effectively when applied to more realistic, imbalanced datasets reflective of actual clinical settings.}

\subsection{Non-linear Behavior of Evaluation Metrics with Temperature Adjustment}
In experiments with different temperature values, it was expected that gradually reducing the temperature would improve both accuracy and F1-Score, particularly for complex prompts with classification examples. The rationale behind this expectation is that a lower temperature should reduce the randomness in responses, leading to more consistent predictions.

However, the results revealed a non-linear behavior. Temperatures between 0.0 and 0.2 achieved the best balance between precision and stability, while higher values, such as 0.3 and above, introduced greater variability, negatively affecting the consistency of predictions. This suggests the existence of an optimal temperature where the model maintains both stability and the necessary sensitivity for the task. The non-linear behavior emphasizes the importance of careful calibration and the need for further studies to understand how variability influences results in complex clinical scenarios, such as depression detection.

\subsection{The Importance of Calibration to Reduce False Negatives in Clinical Diagnostics}

In clinical diagnostics, especially for conditions like depression, minimizing \textbf{false negatives} is critical, as failing to correctly identify a depressed patient may delay treatment and lead to severe mental health consequences. Although high sensitivity is desirable, increasing the model's ability to detect all positive cases can result in a higher number of \textbf{false positives}. However, in healthcare contexts, this trade-off is often acceptable, as it is preferable to investigate more suspected cases than to miss patients who require care.

In our experiments, temperature parameters were selected through a series of exploratory experiments to identify configurations that balance sensitivity and stability. The process involved iteratively testing temperature values within a predefined range (0.0 to 0.5) and analyzing their impact on accuracy, F1-Score, and variability. Lower temperatures demonstrated greater consistency in responses, while higher temperatures introduced randomness, highlighting the trade-offs between predictability and diversity.

% In our experiments,
We observed that the intrinsic variability of the model could affect this calibration. Temperature plays a central role here: reducing randomness yields more consistent and predictable responses but may result in missing some positive cases (false negatives). Conversely, higher temperature values increase variability, potentially improving the model’s sensitivity but also raising the incidence of false positives.

Therefore, careful calibration of the model parameters is essential to balance sensitivity and specificity. In clinical settings, this calibration requires a deep understanding of how variability influences results.

% \textcolor{red}{Future research could focus on automating the parameter selection process by integrating optimization techniques such as Bayesian optimization or grid search. These methods could systematically explore the parameter space, reducing manual intervention and improving reproducibility. Such approaches would enable a more rigorous calibration of parameters, ensuring optimal configurations for clinical applications and further mitigating the effects of variability.}

A potential future approach would be to explore the concept of \textbf{controlled variability}, where multiple runs are performed, and the results are aggregated to increase diagnostic stability. This strategy would help mitigate the impact of inconsistent responses while preserving the model’s ability to correctly identify positive cases. The exploration of this concept could pave the way for automating the calibration process, enabling more efficient and consistent adjustments in clinical settings.

\section{Conclusions and Future Work} 

This preliminary study demonstrates the potential of GPT-4 as a supplementary tool for clinical depression assessment through transcript-based classification. By examining the model's performance across a range of prompt complexities and temperature settings, we have highlighted important insights into how large language models can be optimized for clinical tasks.

The results indicate that the combination of structured prompts with illustrative examples yields promising outcomes, achieving solid performance metrics in both accuracy and F1-Score. Models leveraging complex prompts perform particularly well when temperatures are calibrated between 0.0 and 0.2, ensuring stability and consistency in classification outputs. However, temperatures beyond 0.3 introduced variability, revealing that balancing randomness and stability remains a critical factor in configuring GPT-4 for sensitive clinical use cases.

Overall, our findings suggest that prompt engineering and parameter tuning can greatly enhance the practical utility of GPT-4 in clinical environments. Despite its limitations, this study establishes that GPT-4 can serve as a viable option for mental health assessment, achieving reliable classifications without additional data pre-processing or fine-tuning.

Future advancements in this line of research could further enhance model performance by exploring retrieval-augmented generation (RAG) techniques, expanding the variability analysis of temperature for specific clinical applications such as depression detection, and investigating the role of advanced prompt engineering to optimize outcomes across different scenarios. Additionally, fine-tuning with domain-specific datasets and integrating external medical knowledge bases would help align the model more closely with clinical expectations, fostering AI adoption in healthcare settings.

This study lays a foundation for future work in applying large language models to clinical mental health assessments, demonstrating both the potential and challenges inherent in leveraging state-of-the-art AI tools for patient care.

 \end{document}